\documentclass[]{spie}  

\usepackage{amsmath,amsfonts,amssymb}
\usepackage{graphicx}
\usepackage[multi-part-units=single]{siunitx}
\usepackage[colorlinks=true, allcolors=blue]{hyperref}
\usepackage{todonotes}
\usepackage{pgfplots}
\pgfplotsset{compat=newest}
\usepackage{subcaption}
\usepackage{siunitx}
\title{Two-path 3D CNNs for calibration of system parameters for OCT-based motion compensation}

\author[a]{Nils Gessert$^*$}
\author[a]{Martin Gromniak$^*$}
\author[a]{Matthias Schl\"uter}
\author[a]{Alexander Schlaefer}
\affil[a]{Institute of Medical Technology, Hamburg University of Technology, Am Schwarzenberg-Campus 3, 21073 Hamburg, Geramny}

\authorinfo{Further author information: (Send correspondence to Nils Gessert)\\Nils Gessert: nils.gessert@tuhh.de \\  Martin Gromniak: martin.gromniak@tuhh.de \\ $^*$ Both authors contributed equally.}

\begin{document} 
\maketitle

\begin{abstract}

Automatic motion compensation and adjustment of an intraoperative imaging modality's field of view is a common problem during interventions. Optical coherence tomography (OCT) is an imaging modality which is used in interventions due to its high spatial resolution of few micrometers and its temporal resolution of potentially several hundred volumes per second. However, performing motion compensation with OCT is problematic due to its small field of view which might lead to tracked objects being lost quickly. We propose a novel deep learning-based approach that directly learns input parameters of motors that move the scan area for motion compensation from optical coherence tomography volumes. We design a two-path 3D convolutional neural network (CNN) architecture that takes two volumes with an object to be tracked as its input and predicts the necessary motor input parameters to compensate the object's movement. In this way, we learn the calibration between object movement and system parameters for motion compensation with arbitrary objects. Thus, we avoid error-prone hand-eye calibration and handcrafted feature tracking from classical approaches. We achieve an average correlation coefficient of $0.998$ between predicted and ground-truth motor parameters which leads to sub-voxel accuracy. Furthermore, we show that our deep learning model is real-time capable for use with the system's high volume acquisition frequency.

\end{abstract}


\section{INTRODUCTION}
\label{sec:intro}  
Optical coherence tomography (OCT) is an interferometric imaging modality that allows for volumetric imaging with micrometer-level resolution. OCT has been used in intraoperative scenarios\cite{Lankenau.2007,ehlers2014integrative} such as neurosurgery\cite{Finke.2012} and ophthalmic surgery\cite{Tao.2014}. Recently, systems with high-frequency acquisition have been proposed\cite{NOVAIS201680,siddiqui2018high} which allows for fast imaging and object tracking during 
interventions. As the field of view (FOV) of high-resolution imaging modalities is often limited, the current region-of-interest (ROI) might be lost quickly due to patient and surgical tool movement. As manual adjustment of the imaging system's FOV disrupts the surgical workflow, automatic adjustment is desirable for keeping track of the current ROI. So far, OCT-based tracking and compensation can be performed with markerless approaches using cumbersome and potentially error-prone image-based registration\cite{Laves.2017}. Alternatively, markers can be introduced to the setup which can be invasive but promises higher accuracy. For example, detection of artificial landmarks carved into bone structures has been shown\cite{Zhang.2014b}. More recently, a deep learning-based method has been proposed where a model learns to estimate the pose of a very small, arbitrary marker geometry directly from OCT volumes\cite{gessert2018deep}. For motion compensation, all these methods require a hand-eye calibration between imaging system and compensation device which is difficult for OCT\cite{Rajput.2016}.

\begin{figure}
\centering
\includegraphics[width=0.30\textwidth]{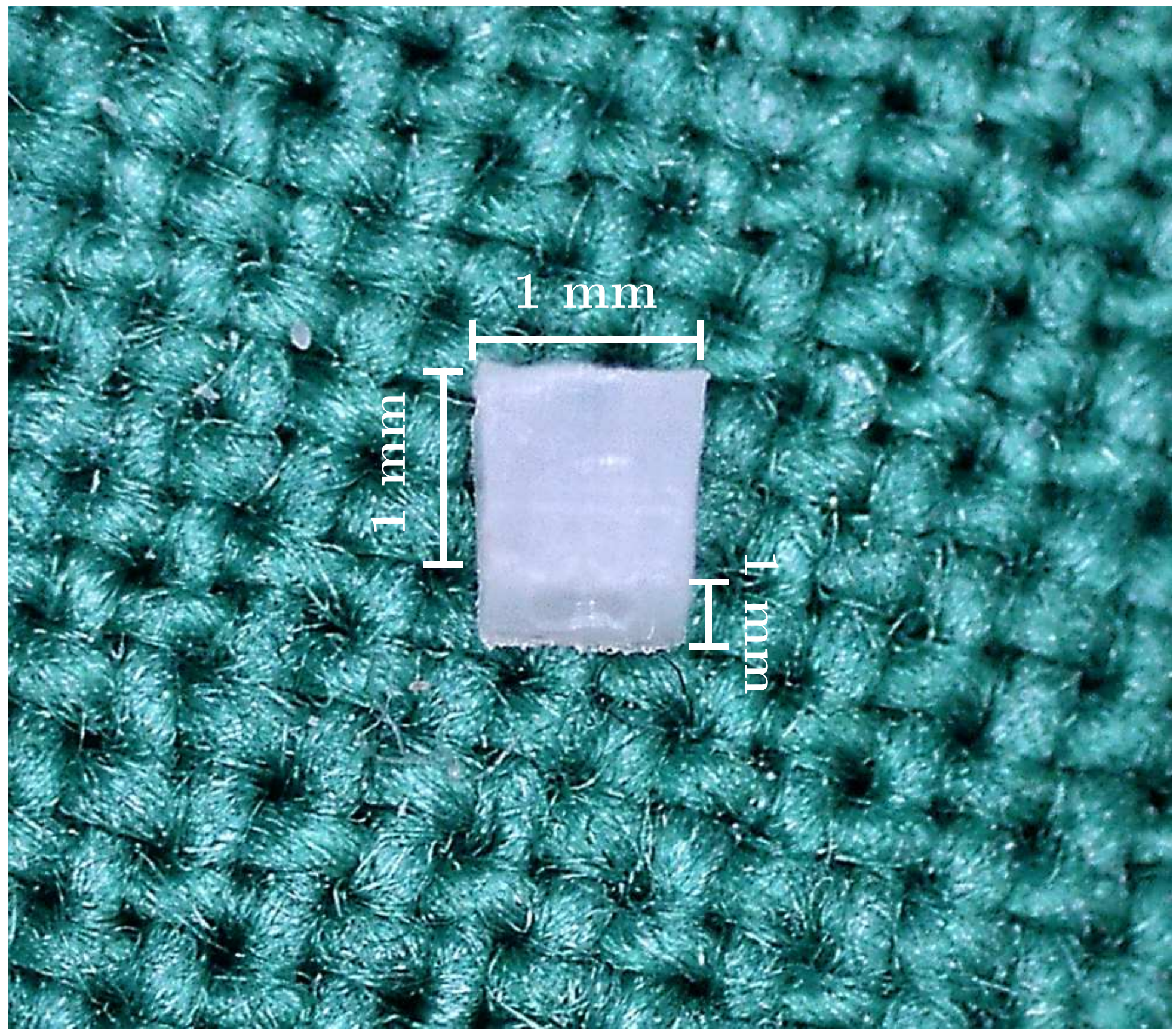}
\includegraphics[width=0.264\textwidth]{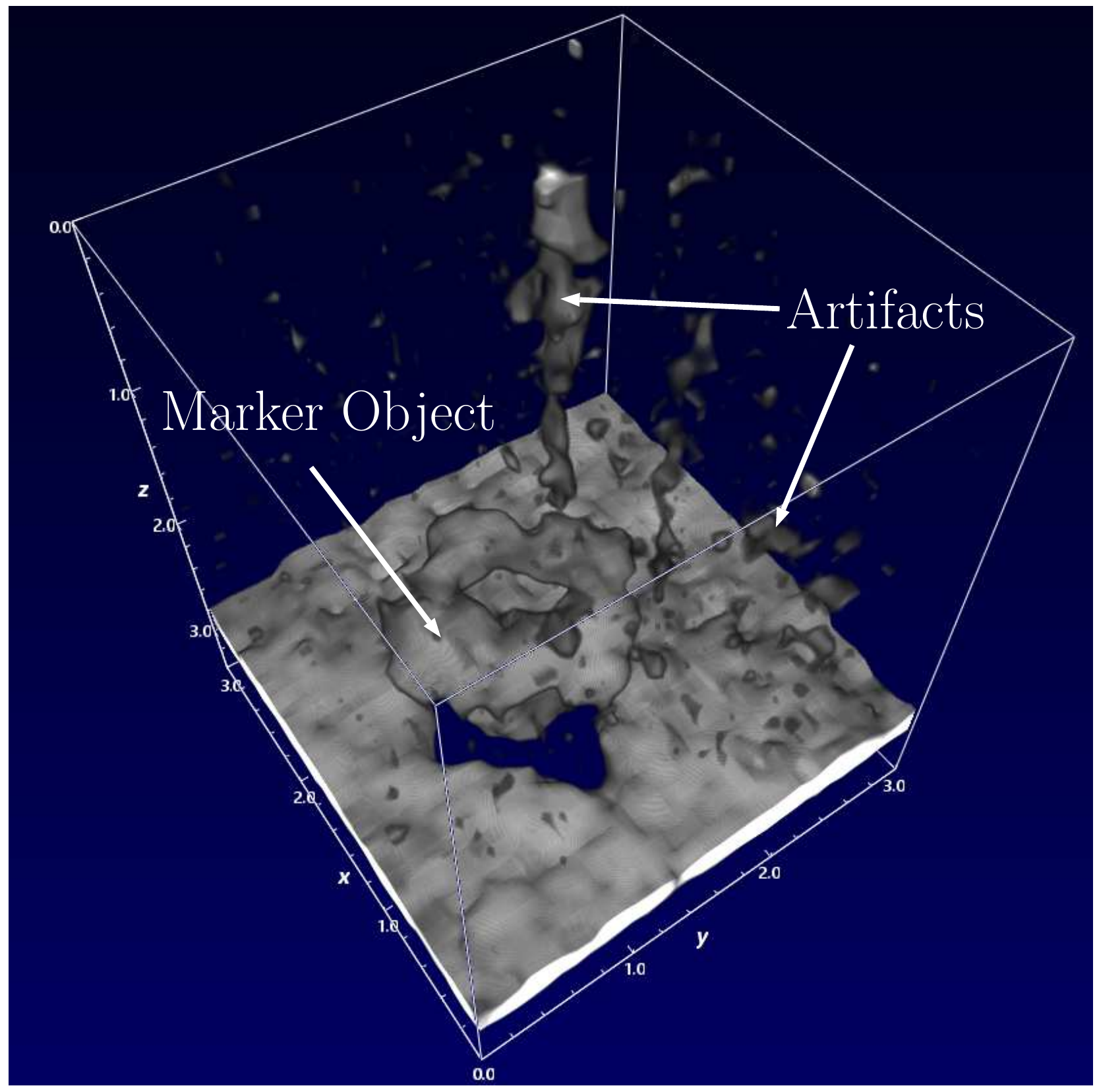}
\includegraphics[width=0.4\textwidth]{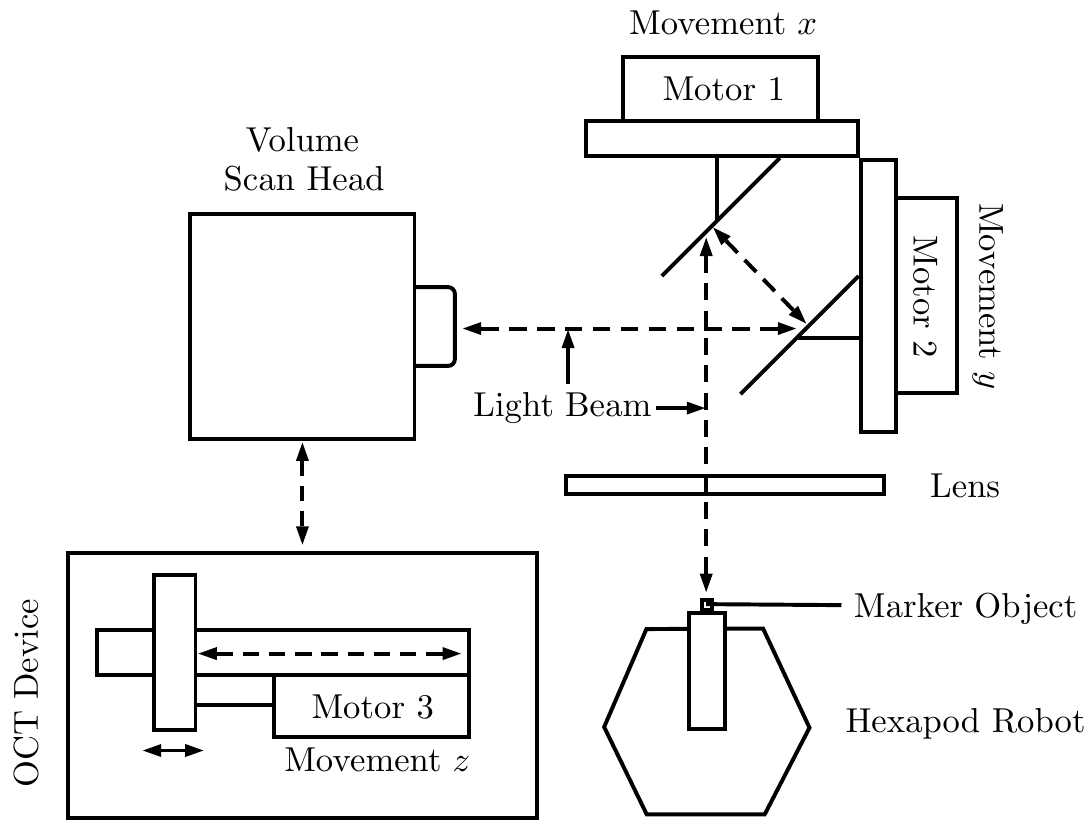}
\caption{The object geometry to be tracked is shown under a digital microscope (left) and in a rendered OCT volume (center). The experimental setup for motion compensation and data acquisition is shown as a draft (right).}
\label{fig:setup}
\end{figure}

In this paper, we propose a calibration strategy between OCT volumes and a compensation system for marker-based tracking. We consider the setup shown in Figure~\ref{fig:setup} with an OCT system that has a mechanism for lateral and axial FOV adjustment. Two motors control mirrors for lateral beam redirection and one motor controls a reference arm for adjusting the scan distance. Thus, the motors can be used to compensate motion of the marker object and keep it within the FOV. In order to compensate motion of the object in this setup, a classic approach first requires an OCT-based hand-eye calibration. Second, either an image-based registration of OCT volumes\cite{niemeijer2009registration} is required or a known marker geometry needs to be detected. Instead, we propose a new direct calibration approach between volumes and motors that combines both steps in a single deep learning model. For this purpose, we extend the recent idea of a 3D convolutional neural network (CNN) model for the estimation of an arbitrary marker's pose\cite{gessert2018deep} to the calibration problem. Instead of a single volume, our model receives two volumes with an object in different areas of the FOV. The two volumes are processed with a two-path 3D CNN architecture \cite{gessert2018force}. At the output, the model predicts motor steps that need to be driven to compensate the motion between the two object states. In this way, we combine hand-eye calibration of OCT volumes with motors and marker detection in a single trainable model. For training of the two-path 3D CNN model we acquire 7 datasets of an object and we show with a separate dataset that the model learns to compensate the object's motion. Thus, the object can be effectively used as marker to keep track of a desired region of interest. Last, we show that the model has low inference times which allows for real-time estimation, despite performing volumetric data processing.


\section{METHODS AND MATERIALS}

\subsection{Experimental Setup} \label{sec:setup}

The setup for OCT-based motion compensation and the object to be tracked is shown are Figure~\ref{fig:setup}. The marker object is milled from a polyoxymethylene block with a size of $\SI{1 x 1 x 1}{\milli \metre}$. We carved out an inner structure in order have subsurface features that can be imaged by OCT and exploited by deep learning models, as recently suggested\cite{gessert2018deep}. The setup itself consists of a swept-source OCT device (OMES, Optores) with an A-Scan rate of $\SI{1.59}{\mega\hertz}$. We use a scan head that provides volumes of size $32\times 32\times 460$ voxels. This leads to a potential acquisition speed of $833$ volumes per second. For a uniform volume size and reduced processing time, we downsample the volumes to a size of $32\times 32\times 32$ voxels which covers a volume of approximately $\SI{3 x 3 x 3.5}{\milli \metre}$. The volume's position in space can be adjusted by three stepper motors. Two motors control mirrors that can laterally move the FOV by \SI{\approx 60}{\milli\metre}. The third motor moves the mirror in the reference arm in a range of \SI{\approx 160}{\milli\metre}. For data acquisition, we also use a hexapod robot with the marker object attached to it. The robot's purpose is to move the object into different orientations for a higher variability in object appearance.

\subsection{Data Acquisition}

For training the deep learning model, a large, labeled dataset is required which we acquire automatically with the setup. In each step, the hexapod moves the object into a random orientation. Then, we move the FOV to two randomly generated motor states $s_1$ and $s_2$ and acquire a volume in each state. Thus, a single labeled example consists of two volumes and the label $s_d = s_1-s_2 = (\Delta x, \Delta y, \Delta z)$ which needs to be driven to overlay the volumes on top of each other. In total, we acquire $7$ datasets with approximately $5000$ examples each. Between each dataset acquisition we rearrange the marker in order to avoid overfitting to a particular initial marker pose.

\subsection{Model} \label{sec:model}

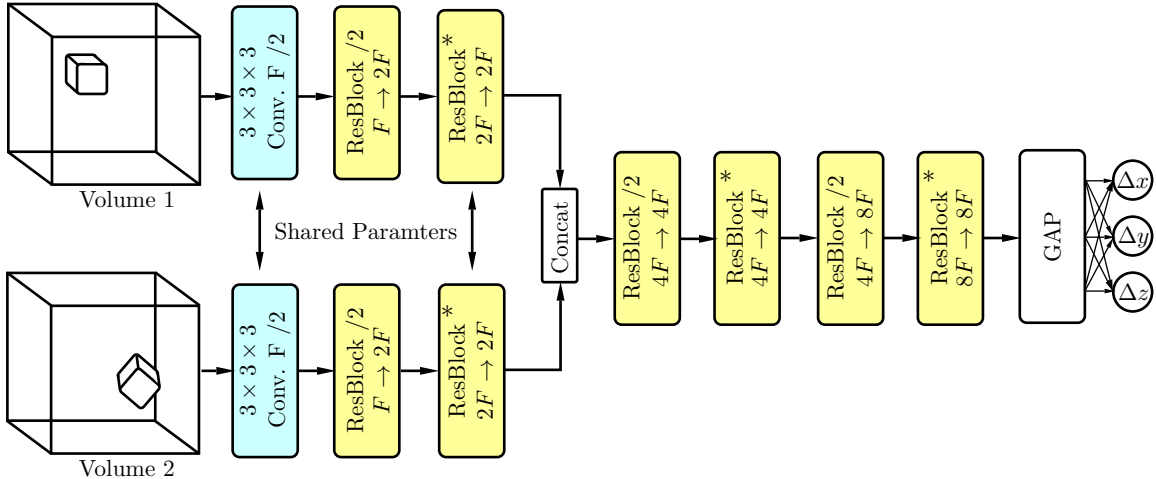
\begin{figure} \centering
\begin{tikzpicture}[      
        every node/.style={anchor=south west,inner sep=0pt},
        x=1mm, y=1mm,
      ]   
     \node (fig1) at (0,0)
       {\includegraphics[angle=90,width=0.9\textwidth]{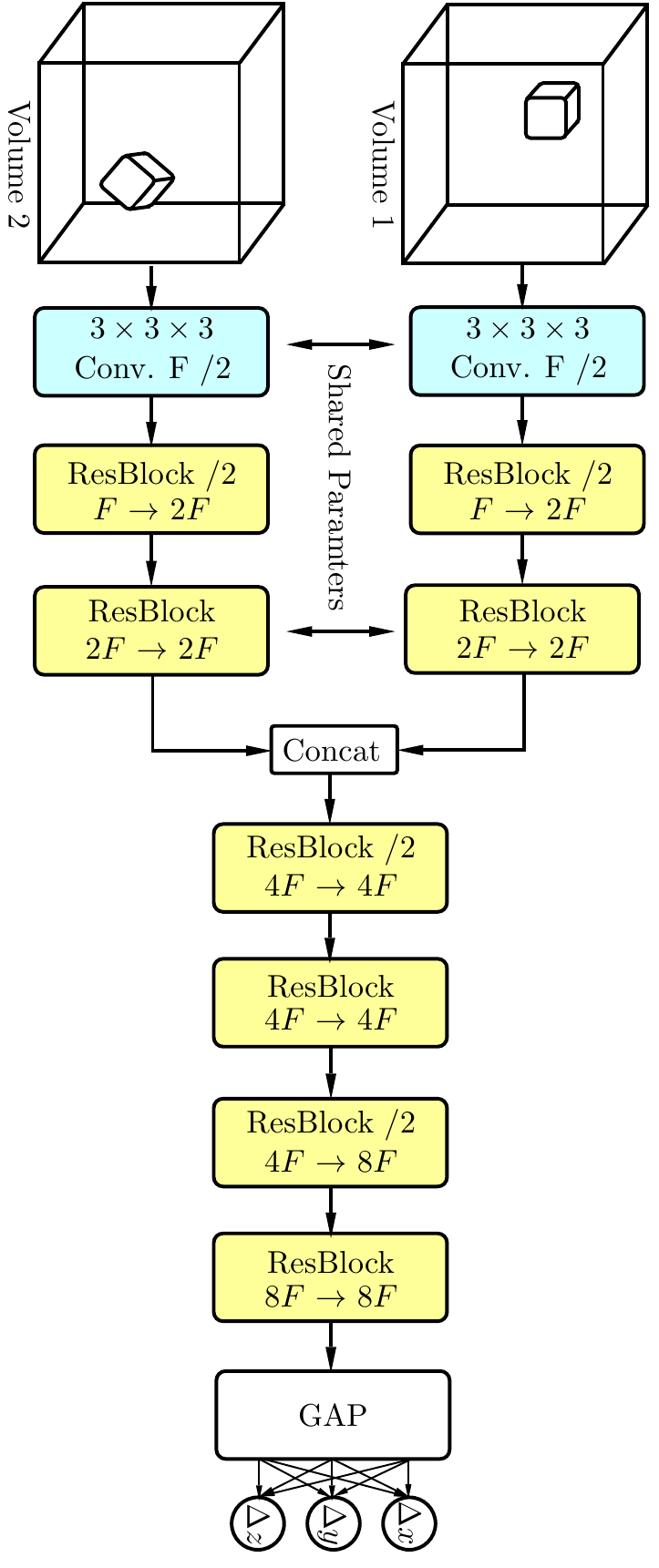}};
       \node[] at (58,57) {*};
       \node[] at (58,21) {*};
       \node[] at (95,39) {*};
       \node[] at (122,39) {*};
\end{tikzpicture}
\caption{The proposed two-path 3D CNN architecture. In each block, the change in the number of feature maps is denoted. $F$ refers to the base number of feature maps. ResBlock refers to residual blocks as introduced by He et al.\cite{He.2016b}. /2 denotes spatial downsampling with a stride of 2. Note, that in the initial two path the model parameters are shared. Concat denotes tensor concatenation along the feature map dimension. GAP denotes global average pooling. GAP is followed by a single linear fully-connected layer. The ResBlocks marked with an asterisk are omitted in the reduced architecture. }
\label{fig:arch}
\end{figure}

The two-path 3D CNN architecture we employ is shown in Figure~\ref{fig:arch}. Each path receives a volume which is processed independently up to a concatenation point. Afterwards, the features are processed jointly and finally, the state difference $s_d$ that would be required for compensation is predicted at the output. We rely on the ResNet principle\cite{He.2016} with identity connections for improved gradient propagation. We also share parameters between the two paths as they receive similar volumes and therefore are likely to require similar features. As the OCT system's volume acquisition rate is very high, we investigate the performance-inference time trade-off by considering downscaled variants of the model shown in Figure~\ref{fig:arch}. For this purpose we vary the base feature map size $F$ which controls the overall capacity of the model. Also, we consider a reduced version of the architecture with less ResBlocks. We train the models with a mean squared error loss using stochastic gradient descent using the Adam\cite{Kingma.2014} optimizer, a constant learning rate of \num{5e-4} and a batch size of $40$.  We split off an entire independent dataset for testing with $5000$ examples. We implement our model using Tensorflow\cite{Abadi.2016}. Training and inference time tests are performed on an NVIDIA GeForce GTX 1080 Ti graphics card.

\section{RESULTS}

\begin{table}
\caption{Performance results on the test set and inference times for the different models. MAE refers to the mean absolute error in motor steps. $2.5$ motor steps in $x$ and $y$ direction roughly correspond to a shift of one voxel. For the $z$ direction, roughly $190$ motor steps correspond to a shift of one voxel. ACC denotes the average correlation coefficient between predictions and targets. $F$ denotes the base number of feature maps, see Figure~\ref{fig:arch}. Red. denotes a reduced architecture with less ResBlocks.}
\label{tab:res}
\centering
\begin{tabular}{l l l l l l l}
	& MAE $\Delta x$ & MAE $\Delta y$ & MAE $\Delta z$ & ACC & & Inf. Time \\ \hline 
    Resnet $F=60$ & \boldmath $1.628 \pm 1.326$ & $1.426 \pm 1.166$ & \boldmath $42.41 \pm 34.34$ & $0.9983$ & & $\SI{7.51 \pm 0.12}{\milli\second}$ \\
    Resnet $F=45$ & $1.990 \pm 1.606$ & $1.634 \pm 1.290$ & $48.06 \pm 36.45$ & \boldmath $0.9984$ & & $\SI{5.40 \pm 0.12}{\milli\second}$  \\    
    Resnet $F=30$ & $1.633 \pm 1.270$ & \boldmath $1.285 \pm 1.078$ & $43.83 \pm 34.84$ & $0.9983$ & & $\SI{3.73 \pm 0.16}{\milli\second}$ \\      
    Resnet $F=15$ & $1.792 \pm 1.393$ & $1.564 \pm 1.243$ & $53.12 \pm 42.13$ & $0.9978$ & & $\SI{2.51 \pm 0.17}{\milli\second}$ \\  
    Resnet $F=15$ Red. & $2.008 \pm 1.619$ & $1.728 \pm 1.405$ & $55.91 \pm 45.84$ & $0.9966$ & & \boldmath $\SI{1.79 \pm 0.16}{\milli\second}$ \\ 
\end{tabular}
\end{table}


The performance results on the test set and inference times for several model variants with differently sized architectures are shown in Table~\ref{tab:res}. Overall, the models' performance is very high with an ACC larger than $0.996$. When downscaling the model, performance slightly deteriorates with base feature maps size below $F=30$. However, the inference time of the smallest model is substantially reduced to $\SI{23.8}{\percent}$ of the largest model's inference time while the ACC barely changes. It is notable that the inference time drops even more when removing ResBlocks for $F=15$ on top of the feature map reduction.

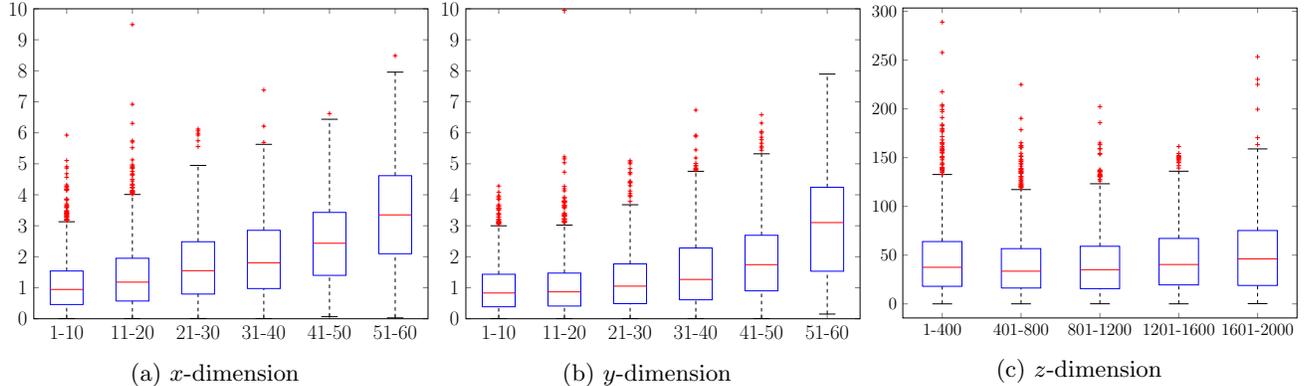
\begin{figure}
\begin{subfigure}[c]{0.33\textwidth}
%
%
\begin{tikzpicture}[scale=0.45]
\LARGE

\begin{axis}[%
width=4.585in,
height=3.608in,
at={(0.769in,0.487in)},
scale only axis,
unbounded coords=jump,
xmin=0.5,
xmax=6.5,
xtick={1,2,3,4,5,6,7},
xticklabels={{1-10},{11-20},{21-30},{31-40},{41-50},{51-60},{}},
ymin=0,
ymax=10,
axis background/.style={fill=white},
legend style={legend cell align=left, align=left, draw=white!15!black}
]
\addplot [color=black, dashed, forget plot]
  table[row sep=crcr]{%
1	1.54075980186462\\
1	3.13017177581787\\
};
\addplot [color=black, dashed, forget plot]
  table[row sep=crcr]{%
2	1.95138359069824\\
2	4.00928497314453\\
};
\addplot [color=black, dashed, forget plot]
  table[row sep=crcr]{%
3	2.48512172698975\\
3	4.94768714904785\\
};
\addplot [color=black, dashed, forget plot]
  table[row sep=crcr]{%
4	2.85332107543945\\
4	5.62733459472656\\
};
\addplot [color=black, dashed, forget plot]
  table[row sep=crcr]{%
5	3.43062591552734\\
5	6.43618011474609\\
};
\addplot [color=black, dashed, forget plot]
  table[row sep=crcr]{%
6	4.6157169342041\\
6	7.95655822753906\\
};
\addplot [color=black, dashed, forget plot]
  table[row sep=crcr]{%
7	8.70961761474609\\
7	8.70961761474609\\
};
\addplot [color=black, dashed, forget plot]
  table[row sep=crcr]{%
1	0.00401163101196289\\
1	0.459331154823303\\
};
\addplot [color=black, dashed, forget plot]
  table[row sep=crcr]{%
2	0.00188636779785156\\
2	0.5721435546875\\
};
\addplot [color=black, dashed, forget plot]
  table[row sep=crcr]{%
3	0.00871467590332031\\
3	0.79718017578125\\
};
\addplot [color=black, dashed, forget plot]
  table[row sep=crcr]{%
4	0.0026092529296875\\
4	0.974693298339844\\
};
\addplot [color=black, dashed, forget plot]
  table[row sep=crcr]{%
5	0.0669212341308594\\
5	1.3996696472168\\
};
\addplot [color=black, dashed, forget plot]
  table[row sep=crcr]{%
6	0.0287437438964844\\
6	2.09793472290039\\
};
\addplot [color=black, dashed, forget plot]
  table[row sep=crcr]{%
7	8.70961761474609\\
7	8.70961761474609\\
};
\addplot [color=black, forget plot]
  table[row sep=crcr]{%
0.875	3.13017177581787\\
1.125	3.13017177581787\\
};
\addplot [color=black, forget plot]
  table[row sep=crcr]{%
1.875	4.00928497314453\\
2.125	4.00928497314453\\
};
\addplot [color=black, forget plot]
  table[row sep=crcr]{%
2.875	4.94768714904785\\
3.125	4.94768714904785\\
};
\addplot [color=black, forget plot]
  table[row sep=crcr]{%
3.875	5.62733459472656\\
4.125	5.62733459472656\\
};
\addplot [color=black, forget plot]
  table[row sep=crcr]{%
4.875	6.43618011474609\\
5.125	6.43618011474609\\
};
\addplot [color=black, forget plot]
  table[row sep=crcr]{%
5.875	7.95655822753906\\
6.125	7.95655822753906\\
};
\addplot [color=black, forget plot]
  table[row sep=crcr]{%
6.875	8.70961761474609\\
7.125	8.70961761474609\\
};
\addplot [color=black, forget plot]
  table[row sep=crcr]{%
0.875	0.00401163101196289\\
1.125	0.00401163101196289\\
};
\addplot [color=black, forget plot]
  table[row sep=crcr]{%
1.875	0.00188636779785156\\
2.125	0.00188636779785156\\
};
\addplot [color=black, forget plot]
  table[row sep=crcr]{%
2.875	0.00871467590332031\\
3.125	0.00871467590332031\\
};
\addplot [color=black, forget plot]
  table[row sep=crcr]{%
3.875	0.0026092529296875\\
4.125	0.0026092529296875\\
};
\addplot [color=black, forget plot]
  table[row sep=crcr]{%
4.875	0.0669212341308594\\
5.125	0.0669212341308594\\
};
\addplot [color=black, forget plot]
  table[row sep=crcr]{%
5.875	0.0287437438964844\\
6.125	0.0287437438964844\\
};
\addplot [color=black, forget plot]
  table[row sep=crcr]{%
6.875	8.70961761474609\\
7.125	8.70961761474609\\
};
\addplot [color=blue, forget plot]
  table[row sep=crcr]{%
0.75	0.459331154823303\\
0.75	1.54075980186462\\
1.25	1.54075980186462\\
1.25	0.459331154823303\\
0.75	0.459331154823303\\
};
\addplot [color=blue, forget plot]
  table[row sep=crcr]{%
1.75	0.5721435546875\\
1.75	1.95138359069824\\
2.25	1.95138359069824\\
2.25	0.5721435546875\\
1.75	0.5721435546875\\
};
\addplot [color=blue, forget plot]
  table[row sep=crcr]{%
2.75	0.79718017578125\\
2.75	2.48512172698975\\
3.25	2.48512172698975\\
3.25	0.79718017578125\\
2.75	0.79718017578125\\
};
\addplot [color=blue, forget plot]
  table[row sep=crcr]{%
3.75	0.974693298339844\\
3.75	2.85332107543945\\
4.25	2.85332107543945\\
4.25	0.974693298339844\\
3.75	0.974693298339844\\
};
\addplot [color=blue, forget plot]
  table[row sep=crcr]{%
4.75	1.3996696472168\\
4.75	3.43062591552734\\
5.25	3.43062591552734\\
5.25	1.3996696472168\\
4.75	1.3996696472168\\
};
\addplot [color=blue, forget plot]
  table[row sep=crcr]{%
5.75	2.09793472290039\\
5.75	4.6157169342041\\
6.25	4.6157169342041\\
6.25	2.09793472290039\\
5.75	2.09793472290039\\
};
\addplot [color=blue, forget plot]
  table[row sep=crcr]{%
6.75	8.70961761474609\\
6.75	8.70961761474609\\
7.25	8.70961761474609\\
7.25	8.70961761474609\\
6.75	8.70961761474609\\
};
\addplot [color=red, forget plot]
  table[row sep=crcr]{%
0.75	0.945616722106934\\
1.25	0.945616722106934\\
};
\addplot [color=red, forget plot]
  table[row sep=crcr]{%
1.75	1.18116188049316\\
2.25	1.18116188049316\\
};
\addplot [color=red, forget plot]
  table[row sep=crcr]{%
2.75	1.55114841461182\\
3.25	1.55114841461182\\
};
\addplot [color=red, forget plot]
  table[row sep=crcr]{%
3.75	1.80629348754883\\
4.25	1.80629348754883\\
};
\addplot [color=red, forget plot]
  table[row sep=crcr]{%
4.75	2.43587875366211\\
5.25	2.43587875366211\\
};
\addplot [color=red, forget plot]
  table[row sep=crcr]{%
5.75	3.35142517089844\\
6.25	3.35142517089844\\
};
\addplot [color=red, forget plot]
  table[row sep=crcr]{%
6.75	8.70961761474609\\
7.25	8.70961761474609\\
};
\addplot [color=black, draw=none, mark=+, mark options={solid, red}, forget plot]
  table[row sep=crcr]{%
1	3.178879737854\\
1	3.18032479286194\\
1	3.18034267425537\\
1	3.18716764450073\\
1	3.18923807144165\\
1	3.19436645507812\\
1	3.1991708278656\\
1	3.21302223205566\\
1	3.26014280319214\\
1	3.27853775024414\\
1	3.28502225875854\\
1	3.29984402656555\\
1	3.32767295837402\\
1	3.33210420608521\\
1	3.35348892211914\\
1	3.38131904602051\\
1	3.3952522277832\\
1	3.39834547042847\\
1	3.40199518203735\\
1	3.41927528381348\\
1	3.44237089157104\\
1	3.45497369766235\\
1	3.48192739486694\\
1	3.48346567153931\\
1	3.58166694641113\\
1	3.59437847137451\\
1	3.63390588760376\\
1	3.66113638877869\\
1	3.70719218254089\\
1	3.8284854888916\\
1	3.85159754753113\\
1	3.86664533615112\\
1	4.14445018768311\\
1	4.16212511062622\\
1	4.18476438522339\\
1	4.1883487701416\\
1	4.24987411499023\\
1	4.27058506011963\\
1	4.31953239440918\\
1	4.56437921524048\\
1	4.67897605895996\\
1	4.86176681518555\\
1	4.90479326248169\\
1	5.10482788085938\\
1	5.92450094223022\\
};
\addplot [color=black, draw=none, mark=+, mark options={solid, red}, forget plot]
  table[row sep=crcr]{%
2	4.0207347869873\\
2	4.04358291625977\\
2	4.06650304794312\\
2	4.08657073974609\\
2	4.08718204498291\\
2	4.09536838531494\\
2	4.11923408508301\\
2	4.12974739074707\\
2	4.13167190551758\\
2	4.22816228866577\\
2	4.23079824447632\\
2	4.23457527160645\\
2	4.29929637908936\\
2	4.31405735015869\\
2	4.34480094909668\\
2	4.4437427520752\\
2	4.49832248687744\\
2	4.6379599571228\\
2	4.64737319946289\\
2	4.67547750473022\\
2	4.75292587280273\\
2	4.84782791137695\\
2	4.89288187026978\\
2	4.89343643188477\\
2	4.93045568466187\\
2	4.96543312072754\\
2	5.12518882751465\\
2	5.51867294311523\\
2	5.70368194580078\\
2	5.73860359191895\\
2	6.30198860168457\\
2	6.9168815612793\\
2	9.49123764038086\\
2	11.2488451004028\\
};
\addplot [color=black, draw=none, mark=+, mark options={solid, red}, forget plot]
  table[row sep=crcr]{%
3	5.5550594329834\\
3	5.7435474395752\\
3	5.92587471008301\\
3	5.98361396789551\\
3	6.05923271179199\\
3	6.11934280395508\\
};
\addplot [color=black, draw=none, mark=+, mark options={solid, red}, forget plot]
  table[row sep=crcr]{%
4	5.68874931335449\\
4	6.21200942993164\\
4	7.37966918945312\\
};
\addplot [color=black, draw=none, mark=+, mark options={solid, red}, forget plot]
  table[row sep=crcr]{%
5	6.62025451660156\\
};
\addplot [color=black, draw=none, mark=+, mark options={solid, red}, forget plot]
  table[row sep=crcr]{%
6	8.48648834228516\\
};
\addplot [color=black, draw=none, mark=+, mark options={solid, red}, forget plot]
  table[row sep=crcr]{%
nan	nan\\
};
\end{axis}
\end{tikzpicture}%
\subcaption{$x$-dimension}
\end{subfigure}
\begin{subfigure}[c]{0.33\textwidth}
%
%
\begin{tikzpicture}[scale=0.45]
\LARGE
\begin{axis}[%
width=4.585in,
height=3.608in,
at={(0.769in,0.487in)},
scale only axis,
unbounded coords=jump,
xmin=0.5,
xmax=6.5,
xtick={1,2,3,4,5,6,7},
xticklabels={{1-10},{11-20},{21-30},{31-40},{41-50},{51-60},{}},
ymin=0,
ymax=10,
axis background/.style={fill=white},
legend style={legend cell align=left, align=left, draw=white!15!black}
]
\addplot [color=black, dashed, forget plot]
  table[row sep=crcr]{%
1	1.43514811992645\\
1	2.99950194358826\\
};
\addplot [color=black, dashed, forget plot]
  table[row sep=crcr]{%
2	1.47316551208496\\
2	3.02116680145264\\
};
\addplot [color=black, dashed, forget plot]
  table[row sep=crcr]{%
3	1.76966142654419\\
3	3.67844581604004\\
};
\addplot [color=black, dashed, forget plot]
  table[row sep=crcr]{%
4	2.28334140777588\\
4	4.75741958618164\\
};
\addplot [color=black, dashed, forget plot]
  table[row sep=crcr]{%
5	2.69306182861328\\
5	5.32147598266602\\
};
\addplot [color=black, dashed, forget plot]
  table[row sep=crcr]{%
6	4.2422924041748\\
6	7.89407730102539\\
};
\addplot [color=black, dashed, forget plot]
  table[row sep=crcr]{%
7	3.86001586914062\\
7	3.86001586914062\\
};
\addplot [color=black, dashed, forget plot]
  table[row sep=crcr]{%
1	0.00028228759765625\\
1	0.38676780462265\\
};
\addplot [color=black, dashed, forget plot]
  table[row sep=crcr]{%
2	0.000787734985351562\\
2	0.410439491271973\\
};
\addplot [color=black, dashed, forget plot]
  table[row sep=crcr]{%
3	0.0033416748046875\\
3	0.48786449432373\\
};
\addplot [color=black, dashed, forget plot]
  table[row sep=crcr]{%
4	0.000949859619140625\\
4	0.611414909362793\\
};
\addplot [color=black, dashed, forget plot]
  table[row sep=crcr]{%
5	0.00302505493164062\\
5	0.900324821472168\\
};
\addplot [color=black, dashed, forget plot]
  table[row sep=crcr]{%
6	0.151992797851562\\
6	1.53362846374512\\
};
\addplot [color=black, dashed, forget plot]
  table[row sep=crcr]{%
7	2.96210098266602\\
7	2.96210098266602\\
};
\addplot [color=black, forget plot]
  table[row sep=crcr]{%
0.875	2.99950194358826\\
1.125	2.99950194358826\\
};
\addplot [color=black, forget plot]
  table[row sep=crcr]{%
1.875	3.02116680145264\\
2.125	3.02116680145264\\
};
\addplot [color=black, forget plot]
  table[row sep=crcr]{%
2.875	3.67844581604004\\
3.125	3.67844581604004\\
};
\addplot [color=black, forget plot]
  table[row sep=crcr]{%
3.875	4.75741958618164\\
4.125	4.75741958618164\\
};
\addplot [color=black, forget plot]
  table[row sep=crcr]{%
4.875	5.32147598266602\\
5.125	5.32147598266602\\
};
\addplot [color=black, forget plot]
  table[row sep=crcr]{%
5.875	7.89407730102539\\
6.125	7.89407730102539\\
};
\addplot [color=black, forget plot]
  table[row sep=crcr]{%
6.875	3.86001586914062\\
7.125	3.86001586914062\\
};
\addplot [color=black, forget plot]
  table[row sep=crcr]{%
0.875	0.00028228759765625\\
1.125	0.00028228759765625\\
};
\addplot [color=black, forget plot]
  table[row sep=crcr]{%
1.875	0.000787734985351562\\
2.125	0.000787734985351562\\
};
\addplot [color=black, forget plot]
  table[row sep=crcr]{%
2.875	0.0033416748046875\\
3.125	0.0033416748046875\\
};
\addplot [color=black, forget plot]
  table[row sep=crcr]{%
3.875	0.000949859619140625\\
4.125	0.000949859619140625\\
};
\addplot [color=black, forget plot]
  table[row sep=crcr]{%
4.875	0.00302505493164062\\
5.125	0.00302505493164062\\
};
\addplot [color=black, forget plot]
  table[row sep=crcr]{%
5.875	0.151992797851562\\
6.125	0.151992797851562\\
};
\addplot [color=black, forget plot]
  table[row sep=crcr]{%
6.875	2.96210098266602\\
7.125	2.96210098266602\\
};
\addplot [color=blue, forget plot]
  table[row sep=crcr]{%
0.75	0.38676780462265\\
0.75	1.43514811992645\\
1.25	1.43514811992645\\
1.25	0.38676780462265\\
0.75	0.38676780462265\\
};
\addplot [color=blue, forget plot]
  table[row sep=crcr]{%
1.75	0.410439491271973\\
1.75	1.47316551208496\\
2.25	1.47316551208496\\
2.25	0.410439491271973\\
1.75	0.410439491271973\\
};
\addplot [color=blue, forget plot]
  table[row sep=crcr]{%
2.75	0.48786449432373\\
2.75	1.76966142654419\\
3.25	1.76966142654419\\
3.25	0.48786449432373\\
2.75	0.48786449432373\\
};
\addplot [color=blue, forget plot]
  table[row sep=crcr]{%
3.75	0.611414909362793\\
3.75	2.28334140777588\\
4.25	2.28334140777588\\
4.25	0.611414909362793\\
3.75	0.611414909362793\\
};
\addplot [color=blue, forget plot]
  table[row sep=crcr]{%
4.75	0.900324821472168\\
4.75	2.69306182861328\\
5.25	2.69306182861328\\
5.25	0.900324821472168\\
4.75	0.900324821472168\\
};
\addplot [color=blue, forget plot]
  table[row sep=crcr]{%
5.75	1.53362846374512\\
5.75	4.2422924041748\\
6.25	4.2422924041748\\
6.25	1.53362846374512\\
5.75	1.53362846374512\\
};
\addplot [color=blue, forget plot]
  table[row sep=crcr]{%
6.75	2.96210098266602\\
6.75	3.86001586914062\\
7.25	3.86001586914062\\
7.25	2.96210098266602\\
6.75	2.96210098266602\\
};
\addplot [color=red, forget plot]
  table[row sep=crcr]{%
0.75	0.830990314483643\\
1.25	0.830990314483643\\
};
\addplot [color=red, forget plot]
  table[row sep=crcr]{%
1.75	0.86762809753418\\
2.25	0.86762809753418\\
};
\addplot [color=red, forget plot]
  table[row sep=crcr]{%
2.75	1.05244255065918\\
3.25	1.05244255065918\\
};
\addplot [color=red, forget plot]
  table[row sep=crcr]{%
3.75	1.26793670654297\\
4.25	1.26793670654297\\
};
\addplot [color=red, forget plot]
  table[row sep=crcr]{%
4.75	1.7454833984375\\
5.25	1.7454833984375\\
};
\addplot [color=red, forget plot]
  table[row sep=crcr]{%
5.75	3.10305595397949\\
6.25	3.10305595397949\\
};
\addplot [color=red, forget plot]
  table[row sep=crcr]{%
6.75	3.41105842590332\\
7.25	3.41105842590332\\
};
\addplot [color=black, draw=none, mark=+, mark options={solid, red}, forget plot]
  table[row sep=crcr]{%
1	3.04242134094238\\
1	3.06608867645264\\
1	3.08211898803711\\
1	3.08924317359924\\
1	3.10763883590698\\
1	3.10944128036499\\
1	3.12128758430481\\
1	3.14462995529175\\
1	3.17272901535034\\
1	3.20220303535461\\
1	3.20696425437927\\
1	3.27808213233948\\
1	3.28701591491699\\
1	3.35221433639526\\
1	3.38819694519043\\
1	3.38955497741699\\
1	3.50846815109253\\
1	3.51925468444824\\
1	3.56014728546143\\
1	3.56445169448853\\
1	3.57350492477417\\
1	3.6227912902832\\
1	3.64195489883423\\
1	3.6790075302124\\
1	3.8479437828064\\
1	3.87050819396973\\
1	3.93031001091003\\
1	3.96533393859863\\
1	4.08513879776001\\
1	4.28109645843506\\
};
\addplot [color=black, draw=none, mark=+, mark options={solid, red}, forget plot]
  table[row sep=crcr]{%
2	3.10010719299316\\
2	3.13024520874023\\
2	3.13082218170166\\
2	3.13512992858887\\
2	3.14929294586182\\
2	3.20216083526611\\
2	3.20696830749512\\
2	3.21271419525146\\
2	3.21858692169189\\
2	3.22633361816406\\
2	3.30156898498535\\
2	3.30744552612305\\
2	3.3250732421875\\
2	3.3386402130127\\
2	3.35874176025391\\
2	3.35954761505127\\
2	3.38767576217651\\
2	3.40146064758301\\
2	3.57549285888672\\
2	3.59262084960938\\
2	3.61786556243896\\
2	3.64846992492676\\
2	3.65095233917236\\
2	3.66868686676025\\
2	3.74227142333984\\
2	3.75053596496582\\
2	3.77787208557129\\
2	3.77864456176758\\
2	3.79070949554443\\
2	3.79801940917969\\
2	3.80545139312744\\
2	3.88055229187012\\
2	3.92183303833008\\
2	4.12841033935547\\
2	4.19624662399292\\
2	4.2734580039978\\
2	4.72970581054688\\
2	5.03345489501953\\
2	5.15975952148438\\
2	5.22211170196533\\
2	9.95177459716797\\
};
\addplot [color=black, draw=none, mark=+, mark options={solid, red}, forget plot]
  table[row sep=crcr]{%
3	3.78567695617676\\
3	3.94432830810547\\
3	3.97632789611816\\
3	4.04180717468262\\
3	4.12155723571777\\
3	4.29098510742188\\
3	4.38428497314453\\
3	4.40095710754395\\
3	4.44003105163574\\
3	4.67419052124023\\
3	4.72303485870361\\
3	4.84402275085449\\
3	5.0023307800293\\
3	5.05477905273438\\
3	5.09835052490234\\
};
\addplot [color=black, draw=none, mark=+, mark options={solid, red}, forget plot]
  table[row sep=crcr]{%
4	4.79589462280273\\
4	4.81336975097656\\
4	4.8292121887207\\
4	4.85080528259277\\
4	4.9005126953125\\
4	4.94740867614746\\
4	5.01414680480957\\
4	5.18741607666016\\
4	5.44783401489258\\
4	5.88736152648926\\
4	5.91606140136719\\
4	6.73146629333496\\
};
\addplot [color=black, draw=none, mark=+, mark options={solid, red}, forget plot]
  table[row sep=crcr]{%
5	5.43363189697266\\
5	5.51091384887695\\
5	5.56595993041992\\
5	5.66322326660156\\
5	5.77386474609375\\
5	5.85581588745117\\
5	5.9931526184082\\
5	6.03907775878906\\
5	6.31204986572266\\
5	6.58051300048828\\
};
\addplot [color=black, draw=none, mark=+, mark options={solid, red}, forget plot]
  table[row sep=crcr]{%
nan	nan\\
};
\addplot [color=black, draw=none, mark=+, mark options={solid, red}, forget plot]
  table[row sep=crcr]{%
nan	nan\\
};
\end{axis}
\end{tikzpicture}%
\subcaption{$y$-dimension}
\end{subfigure}
\begin{subfigure}[c]{0.33\textwidth}
%
%
\begin{tikzpicture}[scale=0.45]
\Large
\begin{axis}[%
width=4.585in,
height=3.608in,
at={(0.769in,0.487in)},
scale only axis,
xmin=0.5,
xmax=5.5,
xtick={1,2,3,4,5},
xticklabels={{1-400},{401-800},{801-1200},{1201-1600},{1601-2000}},
ymin=-14.4431465148926,
ymax=303.482315826416,
axis background/.style={fill=white},
legend style={legend cell align=left, align=left, draw=white!15!black}
]
\addplot [color=black, dashed, forget plot]
  table[row sep=crcr]{%
1	63.9150886535645\\
1	132.709564208984\\
};
\addplot [color=black, dashed, forget plot]
  table[row sep=crcr]{%
2	56.6642456054688\\
2	117.251220703125\\
};
\addplot [color=black, dashed, forget plot]
  table[row sep=crcr]{%
3	59.0992584228516\\
3	123.213134765625\\
};
\addplot [color=black, dashed, forget plot]
  table[row sep=crcr]{%
4	67.1286926269531\\
4	136.068115234375\\
};
\addplot [color=black, dashed, forget plot]
  table[row sep=crcr]{%
5	75.2554626464844\\
5	158.93212890625\\
};
\addplot [color=black, dashed, forget plot]
  table[row sep=crcr]{%
1	0.0080108642578125\\
1	18.0150604248047\\
};
\addplot [color=black, dashed, forget plot]
  table[row sep=crcr]{%
2	0.050323486328125\\
2	16.2640228271484\\
};
\addplot [color=black, dashed, forget plot]
  table[row sep=crcr]{%
3	0.11102294921875\\
3	15.6168670654297\\
};
\addplot [color=black, dashed, forget plot]
  table[row sep=crcr]{%
4	0.044189453125\\
4	19.4645080566406\\
};
\addplot [color=black, dashed, forget plot]
  table[row sep=crcr]{%
5	0.27099609375\\
5	18.895751953125\\
};
\addplot [color=black, forget plot]
  table[row sep=crcr]{%
0.875	132.709564208984\\
1.125	132.709564208984\\
};
\addplot [color=black, forget plot]
  table[row sep=crcr]{%
1.875	117.251220703125\\
2.125	117.251220703125\\
};
\addplot [color=black, forget plot]
  table[row sep=crcr]{%
2.875	123.213134765625\\
3.125	123.213134765625\\
};
\addplot [color=black, forget plot]
  table[row sep=crcr]{%
3.875	136.068115234375\\
4.125	136.068115234375\\
};
\addplot [color=black, forget plot]
  table[row sep=crcr]{%
4.875	158.93212890625\\
5.125	158.93212890625\\
};
\addplot [color=black, forget plot]
  table[row sep=crcr]{%
0.875	0.0080108642578125\\
1.125	0.0080108642578125\\
};
\addplot [color=black, forget plot]
  table[row sep=crcr]{%
1.875	0.050323486328125\\
2.125	0.050323486328125\\
};
\addplot [color=black, forget plot]
  table[row sep=crcr]{%
2.875	0.11102294921875\\
3.125	0.11102294921875\\
};
\addplot [color=black, forget plot]
  table[row sep=crcr]{%
3.875	0.044189453125\\
4.125	0.044189453125\\
};
\addplot [color=black, forget plot]
  table[row sep=crcr]{%
4.875	0.27099609375\\
5.125	0.27099609375\\
};
\addplot [color=blue, forget plot]
  table[row sep=crcr]{%
0.75	18.0150604248047\\
0.75	63.9150886535645\\
1.25	63.9150886535645\\
1.25	18.0150604248047\\
0.75	18.0150604248047\\
};
\addplot [color=blue, forget plot]
  table[row sep=crcr]{%
1.75	16.2640228271484\\
1.75	56.6642456054688\\
2.25	56.6642456054688\\
2.25	16.2640228271484\\
1.75	16.2640228271484\\
};
\addplot [color=blue, forget plot]
  table[row sep=crcr]{%
2.75	15.6168670654297\\
2.75	59.0992584228516\\
3.25	59.0992584228516\\
3.25	15.6168670654297\\
2.75	15.6168670654297\\
};
\addplot [color=blue, forget plot]
  table[row sep=crcr]{%
3.75	19.4645080566406\\
3.75	67.1286926269531\\
4.25	67.1286926269531\\
4.25	19.4645080566406\\
3.75	19.4645080566406\\
};
\addplot [color=blue, forget plot]
  table[row sep=crcr]{%
4.75	18.895751953125\\
4.75	75.2554626464844\\
5.25	75.2554626464844\\
5.25	18.895751953125\\
4.75	18.895751953125\\
};
\addplot [color=red, forget plot]
  table[row sep=crcr]{%
0.75	37.5259437561035\\
1.25	37.5259437561035\\
};
\addplot [color=red, forget plot]
  table[row sep=crcr]{%
1.75	33.531494140625\\
2.25	33.531494140625\\
};
\addplot [color=red, forget plot]
  table[row sep=crcr]{%
2.75	35.0630493164062\\
3.25	35.0630493164062\\
};
\addplot [color=red, forget plot]
  table[row sep=crcr]{%
3.75	40.375244140625\\
4.25	40.375244140625\\
};
\addplot [color=red, forget plot]
  table[row sep=crcr]{%
4.75	46.184814453125\\
5.25	46.184814453125\\
};
\addplot [color=black, draw=none, mark=+, mark options={solid, red}, forget plot]
  table[row sep=crcr]{%
1	133.094360351562\\
1	135.026184082031\\
1	135.521759033203\\
1	137.222412109375\\
1	137.300979614258\\
1	138.145782470703\\
1	138.323486328125\\
1	138.969604492188\\
1	139.789886474609\\
1	139.846923828125\\
1	142.734375\\
1	143.496139526367\\
1	144.533386230469\\
1	147.81623840332\\
1	149.836044311523\\
1	150.108032226562\\
1	151.211883544922\\
1	154.783218383789\\
1	155.468399047852\\
1	155.762680053711\\
1	157.343902587891\\
1	157.621246337891\\
1	158.512908935547\\
1	158.84455871582\\
1	159.880004882812\\
1	161.487152099609\\
1	163.724243164062\\
1	164.956649780273\\
1	165.128753662109\\
1	165.37873840332\\
1	168.105499267578\\
1	171.609771728516\\
1	176.387023925781\\
1	177.987045288086\\
1	178.113433837891\\
1	179.810455322266\\
1	182.629013061523\\
1	183.943756103516\\
1	191.031661987305\\
1	197.198974609375\\
1	199.225067138672\\
1	202.807479858398\\
1	204.046630859375\\
1	217.497421264648\\
1	257.779327392578\\
1	289.031158447266\\
};
\addplot [color=black, draw=none, mark=+, mark options={solid, red}, forget plot]
  table[row sep=crcr]{%
2	119.252807617188\\
2	119.411651611328\\
2	120.498474121094\\
2	120.542663574219\\
2	121.294799804688\\
2	122.083435058594\\
2	122.9228515625\\
2	123.126342773438\\
2	124.285552978516\\
2	125.444396972656\\
2	126.713562011719\\
2	128.172790527344\\
2	129.149536132812\\
2	130.495269775391\\
2	130.710571289062\\
2	133.669525146484\\
2	134.031372070312\\
2	134.672180175781\\
2	135.139007568359\\
2	136.118896484375\\
2	136.24755859375\\
2	136.333740234375\\
2	137.952270507812\\
2	138.410766601562\\
2	138.9560546875\\
2	141.347473144531\\
2	143.415588378906\\
2	144.300109863281\\
2	146.089416503906\\
2	146.631774902344\\
2	148.267639160156\\
2	151.087219238281\\
2	151.723327636719\\
2	154.409790039062\\
2	160.060119628906\\
2	160.56396484375\\
2	161.853759765625\\
2	162.434387207031\\
2	165.211303710938\\
2	178.559906005859\\
2	190.246688842773\\
2	224.911972045898\\
};
\addplot [color=black, draw=none, mark=+, mark options={solid, red}, forget plot]
  table[row sep=crcr]{%
3	126.001586914062\\
3	127.584716796875\\
3	130.337524414062\\
3	130.614929199219\\
3	130.692016601562\\
3	131.322998046875\\
3	131.54931640625\\
3	131.659790039062\\
3	132.953491210938\\
3	133.400756835938\\
3	135.262084960938\\
3	135.771911621094\\
3	135.812927246094\\
3	136.26611328125\\
3	136.868591308594\\
3	137.503234863281\\
3	153.501159667969\\
3	154.182556152344\\
3	159.098876953125\\
3	159.115600585938\\
3	163.25\\
3	165.001342773438\\
3	185.8115234375\\
3	202.246765136719\\
};
\addplot [color=black, draw=none, mark=+, mark options={solid, red}, forget plot]
  table[row sep=crcr]{%
4	139.330078125\\
4	141.716552734375\\
4	144.781982421875\\
4	145.471557617188\\
4	146.809814453125\\
4	148.37939453125\\
4	148.662719726562\\
4	149.659057617188\\
4	150.080322265625\\
4	151.034057617188\\
4	151.293212890625\\
4	152.324829101562\\
4	154.304931640625\\
4	161.428588867188\\
};
\addplot [color=black, draw=none, mark=+, mark options={solid, red}, forget plot]
  table[row sep=crcr]{%
5	163.456176757812\\
5	170.387939453125\\
5	199.576904296875\\
5	225.07177734375\\
5	230.350708007812\\
5	253.410278320312\\
};
\end{axis}
\end{tikzpicture}%
\subcaption{$z$-dimension}
\end{subfigure}
\caption{Absolute errors versus the magnitude of required motor steps along each dimension for $F=30$. The horizontal axes show the absolute motor steps along each dimension while the vertical axes show the absolute errors of the predictions. The error increases only minor until $\mathtt{\sim}40$ steps in lateral $x$- and $y$-dimension and remains almost constant in the $z$-dimension.}
\label{fig:res}
\end{figure}

Moreover, Figure~\ref{fig:res} shows the absolute motor step errors depending on the absolute motor step distances of the labels along each dimension. With larger motor step distances that need to be compensated, the error would be expected to increase. The plot shows that this increase is only minor for a large portion of the motor step range. It can be used as an indication of the motion magnitude that can be expected to have good tracking performance. 

Assuming a rough calibration factor of $\mathtt{\sim}2.5$ for lateral motor steps to voxels and factor of $\mathtt{\sim}190$ for the axial direction, our method qualitatively achieves sub-voxel accuracy. Considering the volume size of $\SI{3 x 3 x 3.5}{\milli \metre}$ with a resolution of $32\times 32\times 32$ voxels, the absolute errors are well below $\SI{100}{\micro \metre}$. 


\section{DISCUSSION AND CONCLUSION}

We propose a new deep learning-based method for OCT-based motion compensation. In particular, we avoid time-consuming and inaccurate volume-based registration with a subsequent hand-eye calibration by directly learning the calibration between marker object movement observed in 3D volumes and motors which move the scan area. For this purpose, we use a two-path 3D CNN architecture that predicts the required motor steps for motion compensation of an object's movement based on two input volumes. Considering the results in Table~\ref{tab:res}, the very high average correlation coefficient shows that the learning problem is well solved. Also, the absolute errors show that we qualitatively achieve sub-voxel accuracy which translates to errors well below $\SI{100}{\micro \metre}$. With Figure~\ref{fig:res} we can show that even large distances to be compensated only lead to a minor increase in error. Thus, our model should be robust even towards rapid and large motion.
As we reattached the object several times in between dataset acquisition, the results indicate that the model is capable of learning to track the object. Thus, the object can be used as marker in a region-of-interest that should be tracked.
When using a more efficient, downscaled architecture, performance is slightly reduced as a trade-off for faster inference. With $\SI{1.79}{\milli \second}$ the model achieves a processing frequency of $559$ volumes per second which is among the same magnitude as the OCT's acquisition frequency. This shows that our model does not constitute a bottleneck in the entire compensation process despite having to perform volumetric data processing. For future work, our calibration strategy could be extended by using different marker objects in order to achive generalization to new marker types. Also, the more challenging motion compensation task of markerless tissue tracking could be addressed. 




\bibliography{egbib.bib} 
\bibliographystyle{spiebib} 

\end{document}